\documentclass{article}

\usepackage[nonatbib, final]{neurips_2021}
\usepackage{float}
\usepackage{chngpage}
\usepackage{xcolor}

\usepackage[utf8]{inputenc} % allow utf-8 input
\usepackage[T1]{fontenc}    % use 8-bit T1 fonts
\usepackage{hyperref}       % hyperlinks
\hypersetup{
    colorlinks=true,
    linkcolor=cyan,
    filecolor=magenta,      
    urlcolor=blue,
    citecolor=blue,
}

\usepackage{url}            % simple URL typesetting
\usepackage{booktabs}       % professional-quality tables
\usepackage{topcapt}        % nice captions to accompany booktabs
\usepackage{amsfonts}       % blackboard math symbols
\usepackage{nicefrac}       % compact symbols for 1/2, etc.
\usepackage{microtype}      % microtypography
\usepackage{lipsum}
\usepackage{graphicx}
\usepackage{amsmath}
\usepackage{multirow}
\usepackage{xspace}
\usepackage{cite}
\usepackage{listings}

\usepackage{color}
\definecolor{deepblue}{rgb}{0,0,0.5}
\definecolor{deepred}{rgb}{0.6,0,0}
\definecolor{deepgreen}{rgb}{0,0.5,0}

\usepackage[nameinlink,capitalise]{cleveref}
\Crefname{figure}{}{}

\usepackage{array}
\newcolumntype{-}{>{\global\let\currentrowstyle\relax}}
\newcolumntype{^}{>{\currentrowstyle}}

\DeclareFixedFont{\ttb}{T1}{txtt}{bx}{n}{10} % for bold
\DeclareFixedFont{\ttm}{T1}{txtt}{m}{n}{10}  % for normal

\newcommand\pythonstyle{\lstset{
language=Python,
basicstyle=\ttm,
otherkeywords={def},             % Add keywords here
keywordstyle=\ttb\color{deepblue},
emph={MyClass,__init__},          % Custom highlighting
emphstyle=\ttb\color{deepred},    % Custom highlighting style
stringstyle=\color{deepgreen},
frame=tb,                         % Any extra options here
showstringspaces=false            % 
}}

\lstnewenvironment{python}[1][]
{
\pythonstyle
\lstset{#1}
}
{}

\newcommand\pythoninline[1]{{\pythonstyle\lstinline!#1!}}

\renewcommand{\vec}[1]{\mathbf{#1}}

\title{Particle Cloud Generation with Message Passing Generative Adversarial Networks}

\newcommand{\kt}{\ensuremath{k_{\mathrm{T}}}\xspace}
\newcommand{\pt}{\ensuremath{p_{\mathrm{T}}}\xspace}

\newcommand{\TeV}{\ensuremath{\,\text{Te\hspace{-.08em}V}}\xspace}

\newcommand{\PYTHIA} {{\textsc{pythia}}\xspace}
\newcommand{\HERWIG} {{\textsc{herwig}}\xspace}

\newcommand{\MADGRAPH} {{\textsc{MADGRAPH}}\xspace}
\newcommand{\MCATNLO} {{\textsc{MCATNLO}}\xspace}
\newcommand{\MGvATNLO}{\MADGRAPH{}5\_a\MCATNLO\xspace}
\newcommand{\jetnet} {{\textsc{JetNet}}\xspace}

\newcommand{\wass}{\ensuremath{W_1}\xspace}
\newcommand{\wassm}{\ensuremath{W_1^\mathrm M}\xspace}
\newcommand{\wassp}{\ensuremath{W_1^\mathrm P}\xspace}
\newcommand{\wassefp}{\ensuremath{W_1^\mathrm{EFP}}\xspace}
\newcommand{\etarel}{\ensuremath{\eta^\mathrm{rel}}\xspace}
\newcommand{\phirel}{\ensuremath{\phi^\mathrm{rel}}\xspace}
\newcommand{\ptrel}{\ensuremath{\pt^\mathrm{rel}}\xspace}

\begin{document}

\author{
  Raghav Kansal, Javier Duarte, Hao Su\\
  University of California San Diego \\
  La Jolla, CA 92093, USA \\
  \And
  Breno Orzari, Thiago Tomei \\
  Universidade Estadual Paulista \\
  S\~{a}o Paulo/SP - CEP 01049-010, Brazil
  \And
  Maurizio Pierini, Mary Touranakou\thanks{Also at National and Kapodistrian University of Athens, Athens, Greece.}\\
  European Organization for Nuclear Research (CERN) \\
  CH-1211 Geneva 23, Switzerland
  \And
  Jean-Roch Vlimant\\
  California Institute of Technology \\
  Pasadena, CA 91125, USA 
  \And
  Dimitrios Gunopulos\\
  National and Kapodistrian University of Athens \\
  Athens 15772, Greece
}

\maketitle

\begin{abstract}

In high energy physics (HEP), jets are collections of correlated particles produced ubiquitously in particle collisions such as those at the CERN Large Hadron Collider (LHC).
Machine learning (ML)-based generative models, such as generative adversarial networks (GANs), have the potential to significantly accelerate LHC jet simulations.
However, despite jets having a natural representation as a set of particles in momentum-space, a.k.a. a particle cloud, 
% to our knowledge 
there exist no generative models applied to such a dataset. 
In this work, we introduce a new particle cloud dataset (JetNet), and apply to it existing point cloud GANs.
Results are evaluated using (1) 1-Wasserstein distances between high- and low-level feature distributions, (2) a newly developed Fr\'{e}chet ParticleNet Distance, and (3) the coverage and (4) minimum matching distance metrics.
Existing GANs are found to be inadequate for physics applications, hence we develop a new message passing GAN (MPGAN), which outperforms existing point cloud GANs on virtually every metric and shows promise for use in HEP.
We propose JetNet as a novel point-cloud-style dataset for the ML community to experiment with, and set MPGAN as a benchmark to improve upon for future generative models. 
Additionally, to facilitate research and improve accessibility and reproducibility in this area, we release the open-source \jetnet Python package with interfaces for particle cloud datasets, implementations for evaluation and loss metrics, and more tools for ML in HEP development.

% \TODO{Add at least the top tagging dataset to jetnet before this paper is public}

\end{abstract}

% keywords can be removed
%\keywords{Deep Learning \and Simulation \and GANs \and MNIST \and Jets}

\section{Introduction}

Over the past decade, machine learning (ML) has become the de facto way to analyze jets, %collimated sprays of particles
collimated high-energy sprays of particles~\cite{Larkoski:2017jix}
% resulting from quarks and gluons 
% produced at high energy~\cite{Larkoski:2017jix} at the CERN Large Hadron Collider (LHC).
produced at the CERN Large Hadron Collider (LHC). 
To apply ML to jets, the most natural representation is a \emph{particle cloud}, a variable-sized set of points in momentum space, whose radiation pattern contains rich information about the underlying physics known as quantum chromodynamics (QCD).
A fundamental question is whether ML algorithms can model this underlying physics and successfully reproduce the rich high- and low-level structure in jets. % without any structural priors.

Answering this question affirmatively has important practical applications.
At the LHC, large simulated data samples of collision events\footnote{An event is a set of observed particles with well-defined momenta, but not generally well-defined positions.} are generated using Monte Carlo (MC) methods in order to translate theoretical predictions into observable distributions, and ultimately perform physics analyses\footnote{See App.~\ref{app:simhep} for further details on the downstream applications of simulations.}.
These samples, numbering in the billions of events, require computationally expensive modeling of the interaction of particles traversing the detector material.
Recently developed generative frameworks in ML such as generative adversarial networks (GANs), if accurate enough, can be used to accelerate this simulation by potentially five orders of magnitude~\cite{lagan}. 

In this work, we advocate for a benchmark jet dataset (JetNet) and propose several physics- and computer-vision-inspired metrics with which the ML community can improve and evaluate generative models in high energy physics (HEP).
To facilitate and encourage research in this area, as well as to make such research more accessible and reproducible, we release interfaces for public particle cloud datasets such as JetNet, implementations for our proposed metrics, and various tools for ML in HEP development in the \jetnet library~\cite{jetnetlib}.
We next apply existing point cloud GANs on JetNet and find the results to be inadequate for physics applications.
Finally, we develop our own message passing GAN (MPGAN), which dramatically improves results on virtually every metric, and propose it as a benchmark on JetNet.

\section{Jets}
\label{sec:jets}

High-energy proton-proton collisions at the LHC produce elementary particles like quarks and gluons, which cannot be isolated due to the QCD property of color confinement~\cite{Ellis:318585}.
These particles continuously radiate or ``split'' into a set of particles, known as a parton shower. 
Eventually they cool to an energy at which they undergo the process of hadronization, where the fundamental particles combine to form more stable hadrons, such as pions and protons. 
The final set of collimated hadrons produced after such a process is referred to as a jet. 

The task of simulating a single jet can be algorithmically defined as inputting an initial particle, which produces the jet, and outputting the final set of particles a.k.a. the jet constituents. 
Typically in HEP the parton shower and hadronization are steps that are simulated sequentially using MC event generators such as \PYTHIA~\cite{pythia} or \HERWIG~\cite{herwig}. 
Simulating either process exactly is not possible because of the complex underlying physics (QCD), and instead these event generators fit simplified physics-inspired stochastic models, such as the Lund string model for hadronization~\cite{lundstringmodel}, to existing data using MC methods.
% These models are then used to generate more data for use in analyses.
The present work can be seen as an extension of this idea, using a simpler, ML-based model, also fitted to data, for generating the jet in one shot, where we are effectively trading the interpretability of the MC methods for the speed of GPU-accelerated ML generators. 

\paragraph{Representations.}
\label{sec:representations}

\begin{figure}[t!]
    \centering
    \includegraphics[width=0.25\textwidth]{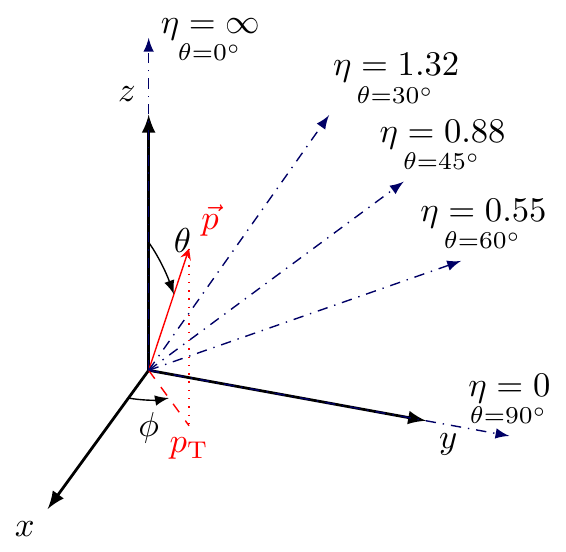}
    \includegraphics[width=0.74\textwidth]{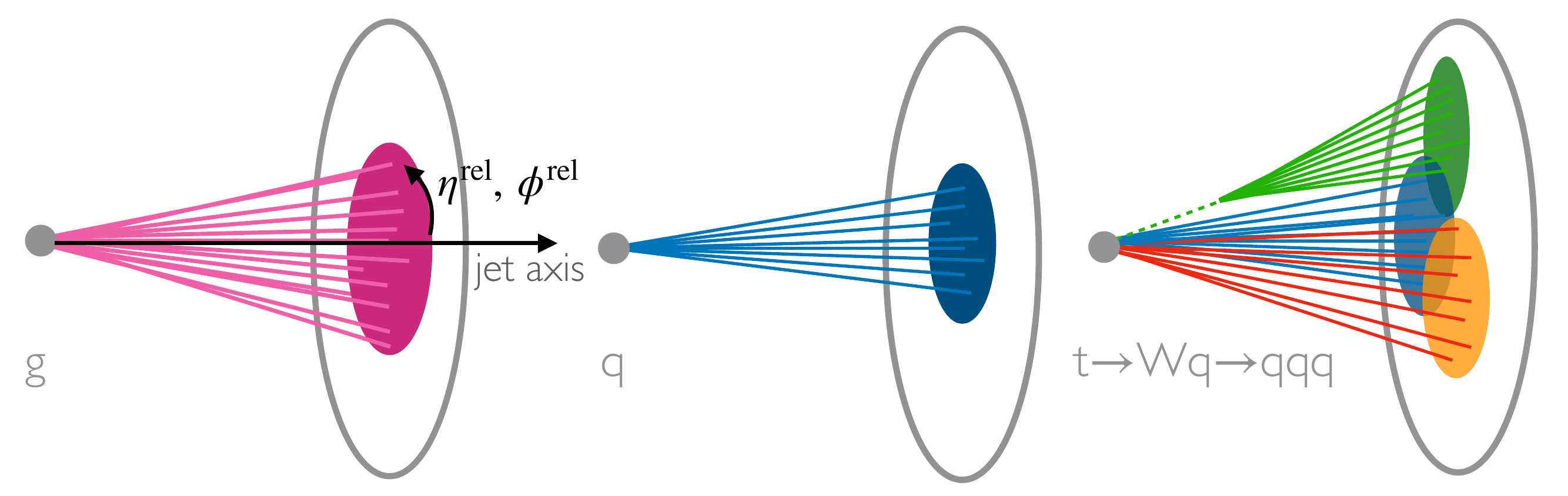}
    \caption{The collider physics coordinate system defining $(\pt, \eta, \phi)$ (left). 
    The three jet classes in our dataset (right). 
    Gluon (g) and light quark (q) jets have simple topologies, with q jets generally containing fewer particles. 
    Top quark (t) jets have a complex three-pronged structure.
    Shown also are the relative angular coordinates $\etarel$ and $\phirel$, measured from the jet axis.}
    \label{fig:jetcartoon}
\end{figure}

As common for collider physics, we use a Cartesian coordinate system with the $z$ axis oriented along the beam axis, the $x$ axis on the horizontal plane, and the $y$ axis oriented upward. 
The $x$ and $y$ axes define the transverse plane, while the $z$ axis identifies the longitudinal direction. 
The azimuthal angle $\phi$ is computed with respect to the $x$ axis. 
The polar angle $\theta$ is used to compute the pseudorapidity $\eta = -\log(\tan(\theta/2))$. 
The transverse momentum ($\pt$) is the projection of the particle momentum on the ($x$, $y$) plane. 
As is customary, we transform the particle momenta from Cartesian coordinates $(p_x, p_y, p_z)$ to longitudinal-boost-invariant pseudo-angular coordinates $(\pt, \eta, \phi)$, as shown in Fig.~\ref{fig:jetcartoon}.

In the context of ML, jets can be represented in multiple ways. 
One popular representation is as images~\cite{Cogan:2014oua,deOliveira:2015xxd}, created by projecting each jet's particle constituents onto a discretized angular $\eta$-$\phi$ plane, and taking the intensity of each ``pixel'' in this grid to be a monotonically increasing function of the corresponding particle $\pt$. 
These tend to be extremely sparse, with typically fewer than 10\% of pixels nonempty~\cite{particlenet}, and the discretization process can furthemore lower the resolution.
% \TODO{Add sample jet image}

Two more spatially efficient representations are as ordered lists or unordered sets~\cite{GNNPP,Duarte:2020ngm} of the jet constituents and their features. 
The difficulty with the former is that there is no particular preferred ordering of the particles---one would have to impose an arbitrary ordering such as by transverse momentum~\cite{Sirunyan:2020lcu}.
The more natural representation is the unordered set of particles in momentum space, which we refer to as a ``particle cloud.'' 
This is in analogy to point cloud representations of 3D objects in position-space prevalent in computer vision created, for example, by sampling from 3D ShapeNet models~\cite{shapenet}. 

Apart from how the samples are produced, significant differences between jets and ShapeNet-based point clouds are that, firstly, jets have physically meaningful low- and high-level features such as particle momentum, total mass of the jet, the number of sub-jets, and $n$-particle energy correlations. 
These physical observables are how we characterize jets, and hence are important to reproduce correctly for physics analysis applications. 
Secondly, unlike the conditional distributions of points given a particular ShapeNet object, which are identical and independent, particle distributions within jets are highly correlated, as the particles each originate from a single source.
The independence of their constituents also means that ShapeNet-sampled point clouds can be chosen to be of a fixed cardinality, whereas this is not possible for jets, which inherently contain varying numbers of particles due to the stochastic nature of particle production.

% to our knowledge all use either a hierarchical sampling approach which is not applicable to jets\footnote{See App.~\ref{pcgen} for further details}, or fixed-size clouds, whereas jets, due to , inherently contain varying numbers of particles.

\paragraph{JetNet.}
\label{sec:dataset}

We publish JetNet~\cite{jetnet} under the CC-BY 4.0 license, to facilitate and advance ML research in HEP, and to offer a new point-cloud-style dataset to experiment with.
Derived from Ref.~\cite{hls4mldata_30p}\footnote{This dataset was also released under the CC-BY 4.0 license.}, it consists of simulated particle jets with transverse momenta $\pt^\mathrm{jet}\approx 1\TeV$, originating from gluons, light quarks, and top quarks produced in 
% $\sqrt{s} =$
$13\TeV$ proton-proton collisions in a simplified detector. 
Technical details of the generation process are given in App.~\ref{app:jetnetgen}.
We limit the number of constituents to the 30 highest $\pt$ particles per jet, allowing for jets with potentially fewer than 30 by zero-padding.
For each particle we provide the following four features: the relative angular coordinates  $\etarel =\eta^\mathrm{particle}
  - \eta^\mathrm{jet}$ and $\phirel =\phi^\mathrm{particle}
  - \phi^\mathrm{jet} \pmod{2\pi}$,  relative transverse momentum
  $\ptrel = \pt^\mathrm{particle}/\pt^\mathrm{jet}$, and a binary mask feature classifying the particle as genuine or zero-padded.

We choose three jet classes, depicted in Fig.~\ref{fig:jetcartoon}, to individually target the unique and challenging properties of jets. 
Gluons provide a useful baseline test, as they typically radiate into a large number of particles before hadronization, largely avoiding the variable-sized cloud issue---at least with a 30 particle maximum, and have a relatively simple topology.
Light quarks share the simple topology, but produce fewer final-state particles, resulting in a larger fraction of zero-padded particles in the dataset. 
They allow evaluation of a model's ability to handle variable-sized clouds.
Finally, top quarks decay into three lighter quarks through an intermediate particle, the W boson, which each may produce their own sub-jets, leading to a complex two- or three-pronged topology---depending on whether the jet clustering algorithm captures all three or just two of these sub-jets.
This results in bimodal jet feature distributions (one peak corresponding to fully merged top quark jets and the other to semi-merged, as seen in Fig.~\ref{fig:results}). 
Thus, top quark jets test models' ability to learn the rich global structure and clustering history of a particle cloud.

\section{Related Work}
\label{sec:relatedwork}

\paragraph{Generative models in HEP.}
\label{sec:genhep}
Past work in this area has exclusively used image-based representations for HEP data.
One benefit of this is the ability to employ convolutional neural network (CNN) based generative models, which have been highly successful on computer vision tasks. 
Refs.~\cite{lagan, calogan, fastcalogan, erdmann2019, carminati2020}, 
% any others to cite?
for example, build upon CNN-based GANs, and Ref.~\cite{sarm} uses an auto-regressive model, to output jet- and detector-data-images. 

In addition to the issues with such representations outlined in Sec.~\ref{sec:representations}, the high sparsity of the images can lead to training difficulties in GANs, and the irregular geometry of the data --- a single LHC detector can typically have multiple sections with differing pixel sizes and shapes --- poses a challenge for CNN GANs which output uniform matrices.
While these can be mitigated to an extent with techniques such as batch normalization~\cite{batchnorm} and using larger/more regular pixels~\cite{fastcalogan}, our approach avoids both issues by generating particle-cloud-representations of the data, as these are inherently sparse data structures and are completely flexible to the underlying geometry.

\paragraph{GANs for point clouds.}
\label{sec:pcgans}

% \TODO{several models built on assumptions not applicable to jets - we go over these in App.~\ref{app:pcgen}.}

% Particle clouds and point clouds have similarities insomuch as they represent sets of elements in some physical space, hence we first test existing point cloud GANs on JetNet. 
There are several published generative models in this area, however the majority exploit inductive biases specific to their respective datasets, such as ShapeNet-based~\cite{pcgan,pointflow,discretepointflow,ShapeGF} and molecular~\cite{kohler20,simm21,gschnet} point clouds, which are not appropriate for jets.
A more detailed discussion, including some experimental results, can be found in App.~\ref{app:pcgen}. 

There do exist some more general-purpose GAN models, namely r-GAN~\cite{rgan}, GraphCNN-GAN~\cite{graphcnngan}, and TreeGAN~\cite{treegan}, and we test these on JetNet.
r-GAN uses a fully-connected (FC) network, GraphCNN-GAN uses graph convolutions based on dynamic $k$-nn graphs in intermediate feature spaces, and TreeGAN iteratively up-samples the graphs with information passing from ancestor to descendant nodes.
In terms of discriminators, past work has used either a FC or a PointNet~\cite{pointnet}-style network.
Ref.~\cite{wang2020rethinking} is the first work to study point cloud discriminator design in detail and finds amongst a number of PointNet and graph convolutional models that PointNet-Mix, which uses both max- and average-pooled features, is the most performant.

We apply the three aforementioned generators and FC and PointNet-Mix discriminators as baselines to our dataset, but find jet structure is not adequately reproduced. 
GraphCNN's local convolutions make learning global structure difficult, and while the TreeGAN and FC generator + PointNet discriminator combinations are improvements, they are not able to learn multi-particle correlations, particularly for the complex top quark jets, nor deal with the variable-sized light quark jets to the extent necessary for physics applications.

\paragraph{Message Passing Neural Networks.}
\label{sec:mpnn}
We attempt to overcome limitations of existing GANs by designing a novel generator and discriminator which can learn such correlations and handle variable-sized particle clouds. 
Both networks build upon the generic message-passing neural network (MPNN)~\cite{MPNN} framework with physics-conscious design choices, and collectively we refer to them as message-passing GAN (MPGAN).
We find MPGAN outperforms existing models on virtually all evaluation metrics.

\subsection{Evaluating generative models.}
\label{sec:geneval}
% \TODO{is subsection ok here? -- Yes, we can put back some subsections if there's space and it makes the paper flow better}

Evaluating generative models is a difficult task, however there has been extensive work in this area in both the physics and computer-vision communities. 

\paragraph{Physics-inspired metrics.}
An accurate jet simulation algorithm should reproduce both low-level and high-level features (such as those described in Sec.~\ref{sec:jets}), hence a standard method of validating generative models, which we employ, is to compare the distributions of such features between the real and generated samples\footnote{More details on the downstream validation procedure used in HEP analyses can be found in App.~\ref{app:simhep}.}~\cite{lagan, calogan, fastcalogan, erdmann2019, carminati2020, fastsim}. 

% \TODO{Define this more formally - maybe in a box like for a theorem?}
For application in HEP, a generative model needs to produce jets with physical features indistinguishable from real.
Therefore, we propose the validation criteria that differences between real and generated sample features may not exceed those between sets of randomly chosen real samples.
To verify this, we use bootstrapping to compare between random samples of only real jets as a baseline. 

A practically useful set of features to validate against are the so-called ``energy-flow polynomials'' (EFPs)~\cite{efps}, which are a set of multi-particle correlation functions. 
Importantly, the set of all EFPs forms a linear basis for all useful jet-level features\footnote{In the context of HEP, this means all infrared- and colinear-safe observables.}.
Therefore, we claim that if we observe all EFP distributions to be reproduced with high fidelity and to match the above criteria, we can conclude with strong confidence that our model is outputting accurate particle clouds. 

\paragraph{Computer-vision-inspired metrics}

A popular metric for evaluating images which has shown to be sensitive to output quality and mode-collapse, though it has its limitations~\cite{borji2021}, is the Fr\'{e}chet Inception Distance~\cite{TTUR} (FID). 
FID is defined as the Fr\'{e}chet distance between Gaussian distributions fitted to the activations of a fully-connected layer of the Inception-v3 image classifier in response to real and generated samples. 
We develop a particle-cloud-analogue of this metric, which we call Fr\'{e}chet ParticleNet Distance (FPND), using the state-of-the-art (SOTA) ParticleNet graph convolutional jet classifier~\cite{particlenet} in lieu of the Inception network.
We note that the FPND and comparing distributions as above is conceptually equivalent, except here instead of physically meaningful and easily interpretable features, we are comparing those found to be statistically optimum for distinguishing jets.

Two common metrics for evaluating point cloud generators are coverage (COV) and minimum matching distance (MMD)~\cite{rgan}. 
Both involve finding the closest point cloud in a sample $X$ to each cloud in another sample $Y$, based on a metric such as the Chamfer distance or the earth mover's distance. 
Coverage is defined as the fraction of samples in $X$ which were matched to one in $Y$, measuring thus the diversity of the samples in $Y$ relative to $X$, and MMD is the average distance between matched samples, measuring the quality of samples. 
We use both, and due to drawbacks of the Chamfer distance pointed out in Ref.~\cite{rgan}, for our distance metric choose only the analogue of the earth mover's distance for particle clouds a.k.a. the energy mover's distance (EMD)~\cite{emd}. 
We discuss the effectiveness and complementarity of all four metrics in evaluating clouds in Sec.~\ref{sec:results}.

% We argue all four metrics are complementary, and necessary to holistically evaluate cloud generation: comparing feature distributions checks whether features necessary for physics analyses are reproduced well; FPND checks if features deemed useful for classification by a SOTA ML classifier are well-reproduced; and MMD and coverage are more focused tests of mode collapse.

\section{MPGAN Architecture}
\label{sec:arch}

\begin{figure}[t!]
    \centering
    \centerline{\includegraphics[width=\textwidth]{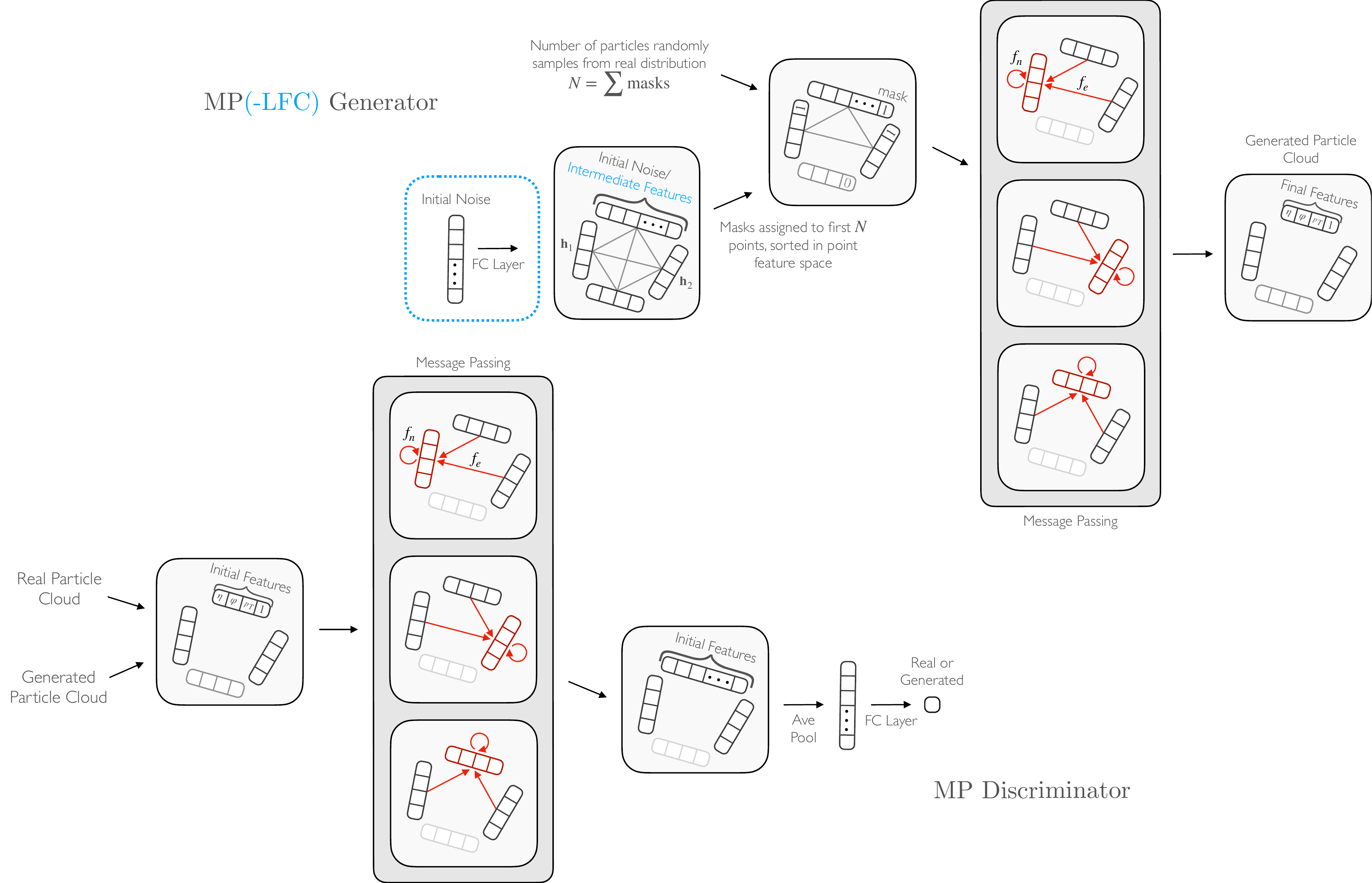}}
    \caption{Top: The MP generator uses message passing to generate a particle cloud. 
    In blue is the initial latent vector and FC layer part of the MP-LFC variant.
    Bottom: The MP discriminator uses message passing to classify an input particle cloud as real or generated.}
    \label{fig:mparch}
\end{figure}

We describe now the architecture of our MPGAN model (Fig.~\ref{fig:mparch}), noting particle cloud-motivated aspects compared to its r-GAN and GraphCNN-GAN predecessors. 
% The main novelties, each elaborated upon below, are (1) the particular message passing operation, (2) a message passing discriminator in addition to the permutation invariant generator, (3) the strategy for generating variable-sized clouds. 

\paragraph{Message passing.}

Jets originate from a single source particle decaying and hadronizing, hence they end up with important high-level jet features and a rich global structure, known as the jet substructure~\cite{Larkoski:2017jix}, stemming from the input particle. 
Indeed any high-level feature useful for analyzing jets, such as jet mass or multi-particle correlations, is necessarily global~\cite{efps}.  
Because of this, while past work in learning on point clouds~\cite{dynamicgraphcnn, monet, particlenet}, including GraphCNN-GAN, has used a locally connected graph structure and convolutions for message passing, we choose a fully connected graph, equally weighting messages from all particles in the clouds. 
Rather than subtracting particle features for messages between particles, useful in graph convolutions to capture local differences within a neighborhood, the respective features are concatenated to preserve the global structure (the difference between particle features is also only physically meaningful if they are in the 4-vector representation of the Lorentz group). 
During the update step in the message passing we find it empirically beneficial to incorporate a residual connection to previous particle features.

The operation can be described as follows. 
For an $N$-particle cloud $J^t = \{p_1^t, \cdots, p_N^t\}$ after $t$ iterations of message passing, with $t=0$ corresponding to the original input cloud, each particle $p^t_i$ is represented by features $\vec{h}^t_i$.
One iteration of message passing is then defined as
\begin{align}
        \label{eqn:mpm} \vec{m}^{t+1}_{ij} &= f_e^{t+1}(\vec{h}^t_i \oplus \vec{h}^t_j), \\
        \label{eqn:mph} \vec{h}^{t+1}_i &= f_n^{t+1}(\vec{h}^t_i \oplus \sum_{j \in J} \vec{m}^{t+1}_{ij})\,,
\end{align}
where $\vec{m}^{t+1}_{ij}$ is the message vector sent from particle $j$ to particle $i$, $\vec{h}^{t+1}_i$ are the updated features of particle $i$, and $f_e^{t+1}$ and $f_n^{t+1}$ are arbitrary functions which, in our case, are implemented as multilayer perceptrons (MLPs) with 3 FC layers.
% Ablation studies using our message passing operation but with locally connected topologies are discussed in App.~\ref{app:locallyconnected}.

\paragraph{Generator.}

We test two initializations of a particle cloud for the MPGAN generator: (1) directly initializing the cloud with $N$ particles with $L$ randomly sampled features, which we refer to as the MP generator, and (2) inputting a single $Z$-dimensional latent noise vector and transforming it via an FC layer into an $N\times L$-dimensional matrix, which we refer to as the MP-Latent-FC (MP-LFC) generator. 
The MP-LFC uses a latent space which can intuitively be understood as representing the initial source particle's features along with parameters to capture the stochasticity of the jet production process.
Due to the complex nature of this process, however, we posit that this global, flattened latent space cannot capture the full phase space of individual particle features. 
Hence, we introduce the MP generator, which samples noise directly per particle, and find that it outperforms MP-LFC (Table~\ref{tab:results}).

% \TODO{can someone confirm that I'm making sense here:}
\paragraph{Discriminator.}

We find the MP generator, in conjunction with a PointNet discriminator, to be a significant improvement on every metric compared to FC and GraphCNN generators.
However, the jet-level features are not yet reproduced to a high enough accuracy (Sec.~\ref{sec:results}).
While PointNet is able to capture global structural information, it can miss the complex interparticle correlations in real particle clouds.
We find we can overcome this limitation by incorporating message passing in the discriminator as well as in the generator.
Concretely, our MP discriminator receives the real or generated cloud and applies MP layers to produce intermediate features for each particle, which are then aggregated via a feature-wise average-pooling operation and passed through an FC layer to output the final scalar feature.
We choose 2 MP layers for both networks. 
% In the generator's case we allow one layer for choosing jet-level and the second for the requisite particle-level features. In the discriminator's case, it's one for learning the jet-level features and the second for evaluating whether they are realistic. 
% \TODO{remove this? feels weak}

\paragraph{Variable-sized clouds.}

In order to handle clouds with varying numbers of particles, as typical of jets, we introduce an additional binary ``masking'' particle feature classifying the particle as genuine or zero-padded. 
Particles in the zero-padded class are ignored entirely in the message passing and pooling operations. 
The MP generator adds mask features to the initial particle cloud, using an additional input of the size of the jet $N$, sampled from the real distribution, before the message passing layers based on sorting in particle feature space. 
Ablation studies with alternative (as well as no) masking strategies are discussed in App.~\ref{app:masking}.

% \TODO{Write masking appendix}

\section{Experiments}
\label{sec:exp}

\paragraph{Evaluation.}
\label{sec:eval}

\begin{table}[tpb]
\centering
\topcaption{\wass distances between real jet mass (\wassm), averaged particle features (\wassp), and averaged jet EFPs (\wassefp) distributions calculated as a baseline, for three classes of jets.
\label{tab:realw1}}
\begin{tabular}{clll}
\toprule
Jet class     & \wassm ($\times 10^{-3}$) & \wassp ($\times 10^{-3}$) & \wassefp ($\times 10^{-5}$) \\
\midrule
Gluon       & $0.7 \pm 0.2$                    & $0.44 \pm 0.09$                   & $0.62 \pm 0.07$                     \\
Light quark & $0.5 \pm 0.1$                     & $0.5 \pm 0.1$                     & $0.46 \pm 0.04$                     \\
Top quark        & $0.51 \pm 0.07$                   & $0.55 \pm 0.07$                   & $1.1 \pm 0.1$                       \\
\bottomrule
\end{tabular}
\end{table}

We use four techniques discussed in Sec.~\ref{sec:geneval} for evaluating and comparing models. 
Distributions of physical particle and jet features are compared visually and quantitatively using the Wasserstein-1 (\wass) distance between them. 
For ease of evaluation, we report (1) the average scores of the three particle features (\wassp) $\etarel$, $\phirel$, and $\ptrel$, (2) the jet mass (\wassm), and (3) the average of a subset of the EFPs\footnote{We choose 5 EFPs corresponding to the set of loopless multigraphs with 4 vertices and 4 edges.}
(\wassefp), which together provide a holistic picture of the low- and high-level aspects of a jet. 
The \wass distances are calculated for each feature between random samples of 10,000 real and generated jets, and averaged over 5 batches. 
Baseline \wass distances are calculated between two sets of randomly sampled real jets with 10,000 samples each, and are listed for each feature in Table \ref{tab:realw1}.
The real samples are split 70/30 for training/evaluation.
We train ParticleNet for classification on our dataset to develop the FPND metric.
FPND is calculated between 50,000 random real and generated samples, based on the activations of the first FC layer in our trained model\footnote{ParticleNet training details are given in App.~\ref{app:pnet}. The trained model is provided in the \jetnet library~\cite{jetnetlib}.}. 
Coverage and MMD are calculated between 100 real and 100 generated samples, and averaged over 10 such batches.
Implementations for all metrics are provided in the \jetnet package~\cite{jetnetlib}.

\paragraph{Results.}
\label{sec:results}

\begin{figure}[htpb]
    \centering
    \centerline{\includegraphics[width=\textwidth]{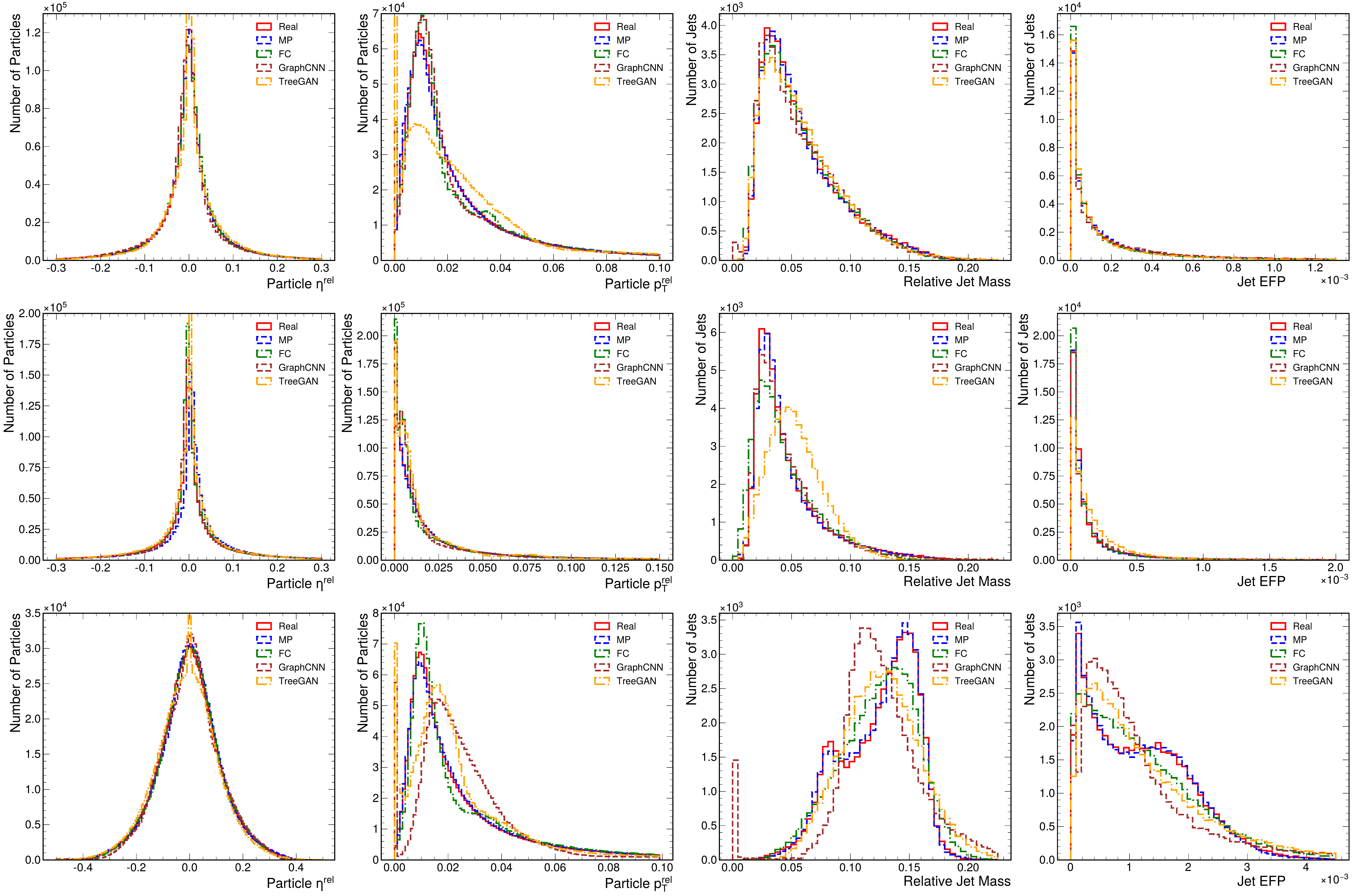}}
    \caption{Comparison of real and generated distributions for a subset of jet and particle features. We use the best performing model for each of the FC, GraphCNN, TreeGAN, and MP generators, as per Table~\ref{tab:results}. Top: gluon jet features, Middle: light quark jets, Bottom: top quark jets.
    % \TODO{plot rgan and graphcnngan with cut-offs for zero-padded}
    }
    \label{fig:results}
\end{figure}

\begin{table}[t]
\topcaption{Six evaluation scores on different generator and discriminator combinations. 
Lower is better for all metrics except COV. 
\label{tab:results}}
\begin{adjustwidth}{-1in}{-1in}% adjust the L and R margins by 1 inch
\centering\resizebox{\textwidth}{!}{
    \begin{tabular}{cllcccccc}
    \toprule
    Jet class & Generator & Discriminator & 
    \begin{tabular}[c]{@{}l@{}}\wassm \\ ($\times 10^{-3}$)\end{tabular} &
    \begin{tabular}[c]{@{}l@{}}\wassp \\ ($\times 10^{-3}$)\end{tabular} &
    \begin{tabular}[c]{@{}l@{}}\wassefp \\ ($\times 10^{-5}$)\end{tabular} & 
    FPND & COV $\uparrow$ & MMD \\
    \midrule
    \multirow{12}{*}{Gluon}  
     & FC & FC & $18.3 \pm 0.2$ & $9.6 \pm 0.4$ & $8.5 \pm 0.5$ & $176$ & $0.24$ & $0.045$\\ 
 & GraphCNN & FC & $2.6 \pm 0.2$ & $9.6 \pm 0.3$ & $12 \pm 8$ & $61$ & $0.39$ & $0.046$\\ 
 & TreeGAN & FC & $41.9 \pm 0.3$ & $69.3 \pm 0.3$ & $14.2 \pm 0.8$ & $355$ & $0.19$ & $0.130$\\ 
 & FC & PointNet & $1.3 \pm 0.4$ & $1.3 \pm 0.2$ & $1.5 \pm 0.9$ & $5.0$ & $0.49$ & $0.039$\\ 
 & GraphCNN & PointNet & $1.9 \pm 0.2$ & $16 \pm 6$ & $200 \pm 1000$ & $7$k & $0.46$ & $0.040$\\ 
 & TreeGAN & PointNet & $1.7 \pm 0.1$ & $4.0 \pm 0.4$ & $4 \pm 1$ & $84$ & $0.37$ & $0.042$\\ 
\cmidrule(lr){2-3}
 & MP & MP & $0.7 \pm 0.2$ & $\mathbf{0.9 \pm 0.3}$ & $\mathbf{0.7 \pm 0.2}$ & $\mathbf{0.12}$ & $\mathbf{0.56}$ & $0.037$\\ 
 & MP-LFC & MP & $\mathbf{0.69 \pm 0.07}$ & $1.8 \pm 0.2$ & $0.9 \pm 0.6$ & $0.20$ & $0.54$ & $0.037$\\ 
\cmidrule(lr){2-3}
 & FC & MP & $4.3 \pm 0.3$ & $21.1 \pm 0.2$ & $9 \pm 1$ & $368$ & $0.11$ & $0.085$\\ 
 & GraphCNN & MP & $2.5 \pm 0.1$ & $9.8 \pm 0.2$ & $13 \pm 8$ & $61$ & $0.38$ & $0.048$\\ 
 & TreeGAN & MP & $2.4 \pm 0.2$ & $12 \pm 7$ & $18 \pm 9$ & $69$ & $0.34$ & $0.048$\\ 
 & MP & FC & $1.2 \pm 0.2$ & $3.7 \pm 0.5$ & $1.6 \pm 0.8$ & $39$ & $0.44$ & $0.040$\\ 
 & MP & PointNet & $1.3 \pm 0.4$ & $1.2 \pm 0.4$ & $4 \pm 2$ & $18$ & $0.53$ & $\mathbf{0.036}$\\ 

    \midrule
    \multirow{12}{*}{\begin{tabular}[c]{@{}l@{}}Light \\ quark\end{tabular}} 
     & FC & FC & $6.0 \pm 0.2$ & $16.3 \pm 0.9$ & $3.9 \pm 0.6$ & $395$ & $0.18$ & $0.053$\\ 
 & GraphCNN & FC & $3.5 \pm 0.2$ & $15.1 \pm 0.4$ & $10 \pm 50$ & $100$ & $0.25$ & $0.038$\\ 
 & TreeGAN & FC & $31.5 \pm 0.3$ & $22.3 \pm 0.4$ & $9.3 \pm 0.4$ & $176$ & $0.06$ & $0.055$\\ 
 & FC & PointNet & $3.1 \pm 0.2$ & $4.5 \pm 0.4$ & $2.3 \pm 0.6$ & $17$ & $0.37$ & $0.028$\\ 
 & GraphCNN & PointNet & $4 \pm 1$ & $5.2 \pm 0.5$ & $50$k $\pm 100$k & $316$ & $0.37$ & $0.031$\\ 
 & TreeGAN & PointNet & $10.1 \pm 0.1$ & $5.7 \pm 0.5$ & $4.1 \pm 0.3$ & $11$ & $0.47$ & $0.031$\\ 
\cmidrule(lr){2-3}
 & MP & MP & $\mathbf{0.6 \pm 0.2}$ & $4.9 \pm 0.5$ & $\mathbf{0.7 \pm 0.4}$ & $0.35$ & $0.50$ & $0.026$\\ 
 & MP-LFC & MP & $0.7 \pm 0.2$ & $\mathbf{2.6 \pm 0.4}$ & $0.9 \pm 0.9$ & $\mathbf{0.08}$ & $\mathbf{0.52}$ & $\mathbf{0.024}$\\ 
\cmidrule(lr){2-3}
 & FC & MP & $6.3 \pm 0.2$ & $16.5 \pm 0.2$ & $4.0 \pm 0.8$ & $212$ & $0.11$ & $0.070$\\ 
 & GraphCNN & MP & $3.5 \pm 0.4$ & $15.0 \pm 0.3$ & $10 \pm 10$ & $99$ & $0.26$ & $0.038$\\ 
 & TreeGAN & MP & $4.8 \pm 0.2$ & $33 \pm 6$ & $10 \pm 2$ & $148$ & $0.22$ & $0.041$\\ 
 & MP & FC & $1.3 \pm 0.1$ & $4.5 \pm 0.4$ & $2.2 \pm 0.6$ & $41$ & $0.37$ & $0.030$\\ 
 & MP & PointNet & $6.5 \pm 0.3$ & $23.2 \pm 0.6$ & $6 \pm 1$ & $850$ & $0.18$ & $0.034$\\ 

    \midrule
    \multirow{12}{*}{\begin{tabular}[c]{@{}l@{}}Top \\ quark\end{tabular}} 
     & FC & FC & $4.8 \pm 0.3$ & $14.5 \pm 0.6$ & $23 \pm 3$ & $160$ & $0.28$ & $0.103$\\ 
 & GraphCNN & FC & $7.0 \pm 0.3$ & $8.0 \pm 0.5$ & $1$k $\pm 6$k & $15$ & $0.48$ & $0.081$\\ 
 & TreeGAN & FC & $17.0 \pm 0.2$ & $19.6 \pm 0.6$ & $33 \pm 2$ & $77$ & $0.39$ & $0.083$\\ 
 & FC & PointNet & $2.7 \pm 0.1$ & $\mathbf{1.6 \pm 0.4}$ & $7.7 \pm 0.5$ & $3.9$ & $0.56$ & $0.075$\\ 
 & GraphCNN & PointNet & $11.3 \pm 0.9$ & $30 \pm 10$ & $37 \pm 2$ & $30$k & $0.39$ & $0.085$\\ 
 & TreeGAN & PointNet & $5.19 \pm 0.08$ & $9.1 \pm 0.3$ & $16 \pm 2$ & $17$ & $0.53$ & $0.079$\\ 
\cmidrule(lr){2-3}
 & MP & MP & $\mathbf{0.6 \pm 0.2}$ & $2.3 \pm 0.3$ & $\mathbf{2 \pm 1}$ & $\mathbf{0.37}$ & $0.57$ & $\mathbf{0.071}$\\ 
 & MP-LFC & MP & $0.9 \pm 0.3$ & $2.2 \pm 0.7$ & $\mathbf{2 \pm 1}$ & $0.93$ & $0.56$ & $0.073$\\ 
\cmidrule(lr){2-3}
 & FC & MP & $6.9 \pm 0.1$ & $39.1 \pm 0.3$ & $15 \pm 1$ & $81$ & $0.26$ & $0.120$\\ 
 & GraphCNN & MP & $6.7 \pm 0.1$ & $8.2 \pm 0.5$ & $40 \pm 10$ & $15$ & $0.49$ & $0.081$\\ 
 & TreeGAN & MP & $13.4 \pm 0.4$ & $45 \pm 7$ & $50 \pm 30$ & $66$ & $0.29$ & $0.101$\\ 
 & MP & FC & $12.9 \pm 0.3$ & $26.3 \pm 0.4$ & $46 \pm 3$ & $58$ & $0.27$ & $0.103$\\ 
 & MP & PointNet & $0.76 \pm 0.08$ & $\mathbf{1.6 \pm 0.4}$ & $4 \pm 1$ & $3.7$ & $\mathbf{0.59}$ & $0.072$\\ 

    \bottomrule
    \end{tabular}
}
\end{adjustwidth}
\end{table}

On each of JetNet's three classes, we test r-GAN's FC, GraphCNN, and TreeGAN generators with rGAN's FC and the PointNet-Mix discriminators, and compare them to MPGAN's MP generator and discriminator models, including both MP and MP-LFC generator variations.
Training and implementation details for each can be found in App.~\ref{app:training}, and all code in Ref.~\cite{mpgancode}. 

We choose model parameters which, during training, yield the lowest \wassm score.
This is because (1) \wass scores between physical features are more relevant for physics applications than the other three metrics, and (2) qualitatively we find it be a better discriminator of model quality than particle features or EFP scores. 
% \TODO{shortcomings of this?} 
Table \ref{tab:results} lists the scores for each model and class, and Fig.~\ref{fig:results} shows plots of selected feature distributions of real and generated jets, for the best performing FC, GraphCNN, TreeGAN, and MP generators.
We also provide in App.~\ref{app:jetimages} discretized images in the angular-coordinates-plane a.k.a ``jet images'', however, we note that it is in general not easy to visually evaluate the quality of individual particle clouds, hence we focus on metrics and visualizations aggregated over batches of clouds.
Overall we find that MPGAN is a significant improvement over the best FC, GraphCNN, and TreeGAN models, particularly for top and light quark jets.
This is evident both visually and quantitatively in every metric, especially jet $\wass$s and FPND, with the exception of \wassp where only the FC generator and PointNet discriminator (FC + PointNet) combination is more performant. 

We additionally perform a latency measurement and find, using an NVIDIA A100 GPU, that MPGAN generation requires 35.7\,$\mu$s per jet. 
In comparison, the traditional generation process for JetNet is measured on an 8-CPU machine as requiring 46ms per jet, meaning MPGAN provides a three-orders-of-magnitude speed-up.
Furthermore, as noted in App.~\ref{app:jetnetgen}, the generation of JetNet is significantly simpler than full simulation and reconstruction used at the LHC, which has been measured to require 12.3s~\cite{pedro19} and 4s~\cite{chen2020data} respectively per top quark jet. 
Hence in practical applications we anticipate MPGAN's improvement to potentially rise to five-orders-of-magnitude.

\paragraph{Real baseline comparison. }

We find that MPGAN's jet-level \wass scores all fall within error of the baselines in Table~\ref{tab:realw1}, while those of alternative generators are several standard deviations away. 
% best performing fc gen is ~4x worse on top jet features
This is particularly an issue with complex top quark particle clouds, where we can see in Fig.~\ref{fig:results} none of the existing generators are able to learn the bimodal jet feature distributions, and smaller light quark clouds, where we see distortion of jet features due to difficulty reproducing the zero-padded particle features. 
No model is able to achieve particle-level scores close to the baseline, and only those of the FC + PointNet combination and MPGAN are of the same order of magnitude.
We conclude that MPGAN reproduces the physical observable distributions to the highest degree of accuracy, but note, however, that it requires further improvement in particle feature reconstruction before it is ready for practical application in HEP.

\paragraph{Architecture discussion. }

To disentangle the effectiveness of the MP generator and discriminator, we train each individually with alternative counterparts (Table~\ref{tab:results}). 
With the same PointNet discriminator, the GraphCNN and TreeGAN generators perform worse than the simple FC generator for every metric on all three datasets.
The physics-motivated MP generator on the other hand outperforms all on the gluon and top quark datasets, and significantly so on the jet-level \wass scores and the FPND. 
We note, however, that the MP generator is not a significant improvement over the other generators with an FC discriminator. 
Holding the generator fixed, the PointNet discriminator performs significantly better over the FC for all metrics.
With the FC, GraphCNN, and TreeGAN generators, PointNet is also an improvement over the MP discriminator.
With an MP generator, the MP discrimimator  is more performant on jet-level \wass and FPND scores but, on the top quark dataset, degrades \wassp relative to PointNet.

We learn from these three things: (1) a generator or discriminator architecture is only as effective as its counterpart---even though the MPGAN combination is the best overall, when paired with a network which is not able to learn complex substructure, or which breaks the permutation symmetry, neither the generator or discriminator is performant, (2) for high-fidelity jet feature reconstruction, both networks must be able to learn complex multi-particle correlations---however, this can come at the cost of low-level feature accuracy, and (3) MPGAN's masking strategy is highly effective as both MP networks are improvements all around on light quark jets.

% The MP generator with an FC discriminator is an improvement over the GraphCNN. This is evidence of MPGAN's fully connected message passing operation being better suited for learning particle clouds and jet structure than graph convolutions. We note, however, that neither MP generator or discriminator alone is sufficient to reproduce the complex top jet topology \TODO{MP disc and MP gen separate top jet plots showing it can't learn complex structure without both}, highlighting the importance of both networks being able to learn jet structure. \TODO{MP gen vs FC/GraphCNN gen light quark jet plots showing zero-padding is throwing it off}. 

\paragraph{Particle cloud evaluation metrics. }

\begin{figure}[htpb]
    \centering
    \centerline{
    \includegraphics[width=\textwidth]{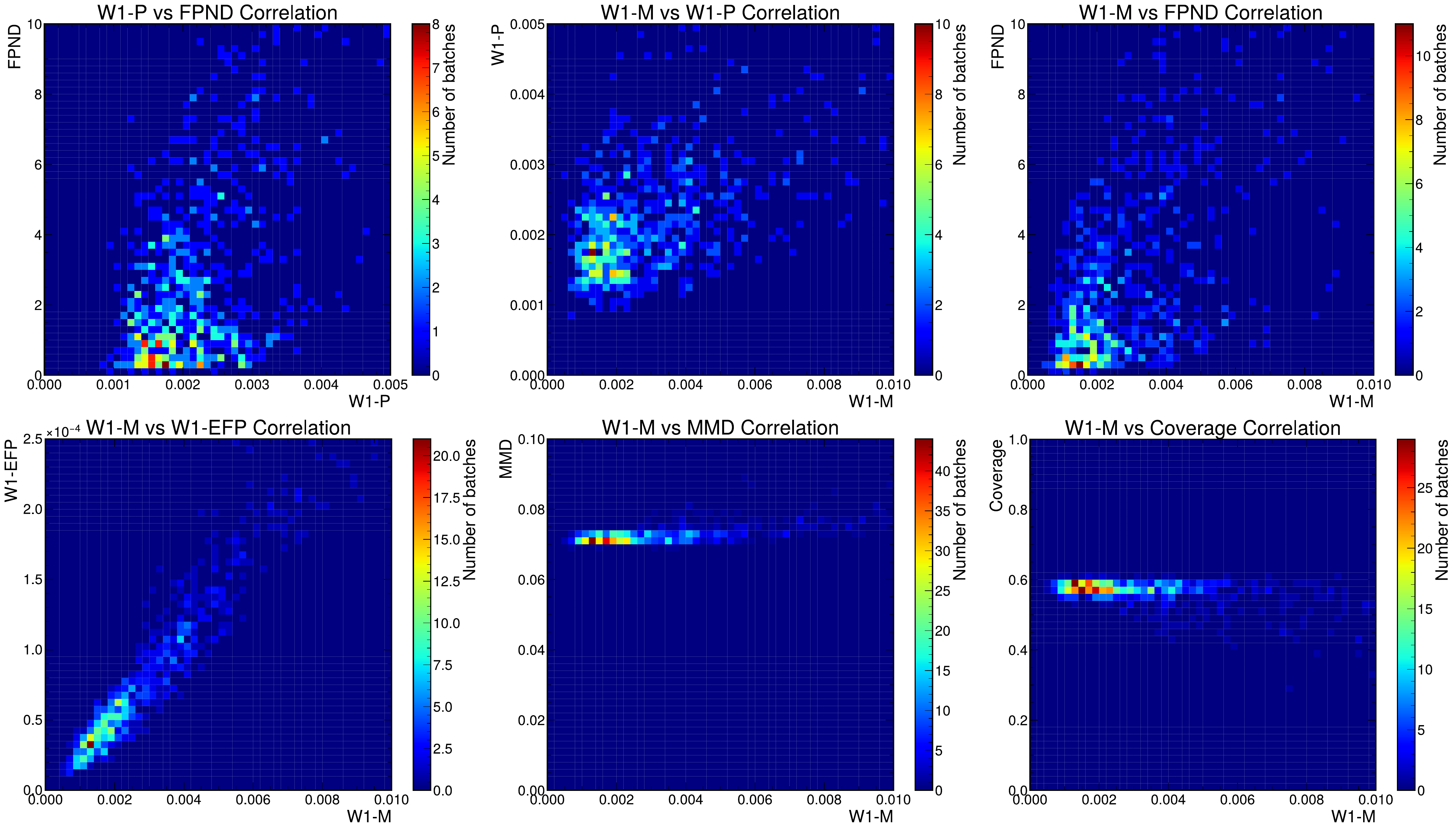}}
    \caption{Correlation plots between pairs of evaluation metrics, evaluated on 400 separate batches of 50,000 MPGAN generated top quark jets.
    }
    \label{fig:correlation}
\end{figure}

We now discuss the merits of each evaluation metrics and provide suggestions for their use in future work. Fig.~\ref{fig:correlation} shows correlation plots between chosen pairs of our evaluation metrics.
As expected, we find W1-M and W1-EFP to be highly correlated,  as they both measure learning of global jet features. 
For rigorous validation we suggest measuring both but for time-sensitive use-cases, such as quick evaluations during model training, W1-M should be sufficient. 
W1-M, FPND, and W1-P are all measuring different aspects of the generation and are relatively uncorrelated. 
We expect FPND overall to be the best and most discriminatory metric for evaluation, as it compares features found by a SOTA classifier to be statistically optimum for characterizing jets, while the W1 scores are valuable for their interpretability. 
Out of these, W1-M/W1-EFP are the most important from a physics-standpoint, as we generally characterize collisions by the high-level features of the output jets, rather than the individual particle features. 

MMD and coverage are both valuable for specifically evaluating the quality and diversity of samples respectively, however we see from Fig.~\ref{fig:correlation} that they saturate after a certain point, after which FPND and \wass scores are necessary for stronger discrimination. 
We also note that in Table~\ref{tab:results}, models with low \wass scores relative to the baseline have the best coverage and MMD scores as well.
This indicates that the \wass metrics are sensitive to both mode collapse (measured by coverage), which is expected as in terms of feature distributions mode collapse manifests as differing supports, to which the \wass distance is sensitive, as well as to individual sample quality (measured by MMD), which supports our claim that recovering jet feature distributions implies accurate learning of individual cloud structure.
Together this suggests that low \wass scores are able validate sample quality and against mode collapse, and justifies our criteria that a practical ML simulation alternative have \wass scores close to the baselines in Table~\ref{tab:results}. 
In conclusion, for thorough validation of generated particle clouds, we recommend considering all three W-1 scores in conjunction with FPND, while MMD and coverage, being focused tests of these aspects of generation, may be useful for understanding failure modes during model development. 

% Each metric proposed here has unique merit. 
% We see that models with low \wass scores relative to the baseline have the best coverage and MMD scores as well.
% This indicates that the \wass metrics are sensitive to both mode collapse (measured by coverage) --- this is expected as, in terms of feature distributions, mode collapse manifests as differing supports, to which the \wass distance is reasonably sensitive, as well as to individual sample quality (measured by MMD) --- this supports our claim that recovering jet feature distributions implies accurate learning of individual cloud structure.
% Together this suggests that low \wass scores are alone sufficient to validate sample quality and against mode collapse, and justifies our criteria that a practical ML simulation alternative have \wass scores close to the baselines in Table~\ref{tab:results}. 
% However, MMD and coverage, being focused tests of these aspects of generation, are useful for understanding failure modes. 

% Additionally, we observe that models with the lowest FPND scores on each dataset also possess the lowest \wass scores, as ParticleNet's deep activations are likely to be correlated with jet features. 
% Overall, we expect FPND to be a better measure of quality since the features being compared are statistically optimum for classifying jets. 
% However, as these features are not necessarily meaningful and less interpretable from a physics-perspective, \wass scores are also valuable.  

\section{Summary}
\label{sec:summary}

In this work, we publish JetNet: a novel particle cloud dataset to advance machine learning (ML) research in high energy physics (HEP), and provide a novel point-cloud-style dataset containing rich underlying physics for the ML community to experiment with.
We apply existing state-of-the-art point cloud generative models to JetNet, and propose several physics- and computer-vision-inspired metrics to rigorously evaluate generated clouds.  
We find that existing models are not performant on a number of metrics, and fail to reproduce high-level jet features---arguably the most significant aspect for HEP.
Our new message-passing generative adversarial network (MPGAN) model, designed to capture complex global structure and handle variable-sized clouds significantly improves performance in this area, as well as other metrics. 
We propose MPGAN as a new baseline model on JetNet and invite others to improve upon it. 

\paragraph{Impact}
\label{sec:impacts}

With our JetNet dataset and library, we hope to lower the barrier to entry, improve reproducibility, and encourage development in HEP and ML. 
Particularly so in the area of simulation, where an accurate and fast ML particle cloud generator will have significant impact in (1) lowering the computational and energy cost of HEP research, as well as (2) increasing precision and sensitivity to new physics at the Large Hadron Collider and future colliders by providing more high-quality simulated data samples.
One negative consequence of this, however, may be a loss of interpretability, and hence trustability, of the particle production generative model, which may ultimately increase uncertainties---though the metrics we propose should mitigate against this.
More broadly, further advancements in the field of ML point cloud generation may result in fake visual data generation for proliferation of misinformation and impersonation/identity theft.

%%% comment out for blind review %%%
\begin{ack}
% \section*{Acknowledgements}
This work was supported by the European Research Council (ERC) under the European Union's Horizon 2020 research and innovation program (Grant Agreement No. 772369).
%%IRIS-HEP fellowship acknowledgment
R.~K. was partially supported by an IRIS-HEP fellowship through the U.S. National Science Foundation (NSF) under Cooperative Agreement OAC-1836650, and by the LHC Physics Center at Fermi National Accelerator Laboratory, managed and operated by Fermi Research Alliance, LLC under Contract No. DE-AC02-07CH11359 with the U.S. Department of Energy (DOE).
J.~D. is supported by the DOE, Office of Science, Office of High Energy Physics Early Career Research program under Award No. DE-SC0021187 and by the DOE, Office of Advanced Scientific Computing Research under Award No. DE-SC0021396 (FAIR4HEP).
B.~O and T.~T are supported by grant 2018/25225-9, S\~{a}o Paulo Research Foundation (FAPESP).
B.~O was also partially supported by grants \#2018/01398-1 and \#2019/16401-0, S\~{a}o Paulo Research Foundation (FAPESP).
J-R.~V. is partially supported by the ERC under the European Union's Horizon 2020 research and innovation program (Grant Agreement No. 772369) and by the DOE, Office of Science, Office of High Energy Physics under Award No. DE-SC0011925, DE-SC0019227, and DE-AC02-07CH11359.
D.~G. is partially supported by the EU ICT-48 2020 project TAILOR (No. 952215).
%% PRP Nautilus acknowledgement:
This work was performed using the Pacific Research Platform Nautilus HyperCluster supported by NSF awards CNS-1730158, ACI-1540112, ACI-1541349, OAC-1826967, the University of California Office of the President, and the University of California San Diego's California Institute for Telecommunications and Information Technology/Qualcomm Institute. 
Thanks to CENIC for the 100\,Gpbs networks.
Funding for cloud credits was supported by NSF Award \#1904444  Internet2 supported E-CAS Exploring Clouds to Accelerate Science.
\end{ack}

%\clearpage
\bibliographystyle{cms_unsrt}
\bibliography{bibliography}

% \clearpage
% \include{checklist}

% \clearpage

\appendix

\section{Simulation-Based Inference in HEP}
\label{app:simhep}

The primary motivation for producing simulations with Monte Carlo (MC) generators, and potentially by machine learning (ML) generators as well, in experimental high energy physics is to develop likelihood models for new fundamental physics theories, with unknown values of parameters of interest~\cite{Cranmer30055,CowanPDGStat}. 
These models are compared with experimental data, such as that collected at the LHC, to perform hypothesis tests of the theories as well as estimate and develop limits and confidence intervals for physical parameters~\cite{CowanPDGStat,Khachatryan:2014jba}.
Increasing the number of simulations and their accuracy can reduce the statistical uncertainties in our models, thus allowing for higher precision measurements and potential discovery of new physics~\cite{Barlow:1993dm,Brehmer:2020cvb}. 

In practice, in each analysis we perform rigorous checks of our simulations, including comparisons in “control regions” (selected portions of the data with a known composition) to check for MC discrepancies with real data. 
Mismatches are corrected via reweighting the events, i.e. unrealistic simulated jets will be given less weight in the overall analysis, and are factored into the final uncertainties in the analysis results. 
The same procedure should be followed for data created by generative ML methods, however, validation before this step using the metrics proposed in Sec.~\ref{sec:relatedwork}, particularly the \wass scores, should mitigate the possibilities of such outliers. 

\section{JetNet Generation}
\label{app:jetnetgen}

The so-called parton-level events are first produced at leading-order using \MGvATNLO2.3.1~\cite{Alwall:2014hca} with the NNPDF\,2.3LO1 parton distribution functions~\cite{Ball:2012cx}.
To focus on a relatively narrow kinematic range, the transverse momenta of the partons and undecayed gauge bosons are generated in a window with energy spread given by $\Delta \pt / \pt = 0.01$, centered at $1\TeV$.
These parton-level events are then decayed and showered in \PYTHIA8.212~\cite{pythia} with the Monash 2013 tune~\cite{Skands:2014pea}, including the contribution from the underlying event.
For each original particle type, 200,000 events are generated. 
Jets are clustered using the anti-$\kt$ algorithm~\cite{Cacciari:2008gp}, with a distance parameter of $R = 0.8$ using the \textsc{FastJet}~3.1.3 and \textsc{FastJet~contrib}~1.027 packages~\cite{fastjet:1,fastjet:2}.
Even though the parton-level $\pt$ distribution is narrow, the jet $\pt$ spectrum is significantly broadened by kinematic recoil from the parton shower and energy migration in and out of the jet cone.
We apply a restriction on the measured jet $\pt$ to remove extreme events outside of a window of $0.8\TeV < \pt < 1.6\TeV$ for the $\pt = 1\TeV$ bin.
This generation is a significantly simplified version of the official simulation and reconstruction steps used for real detectors at the LHC, so as to remain experiment-independent and allow public access to the dataset.

\section{Point Cloud Generative Models}
\label{app:pcgen}

\begin{figure}[htpb]
    \centering
    \centerline{\includegraphics[width=\textwidth]{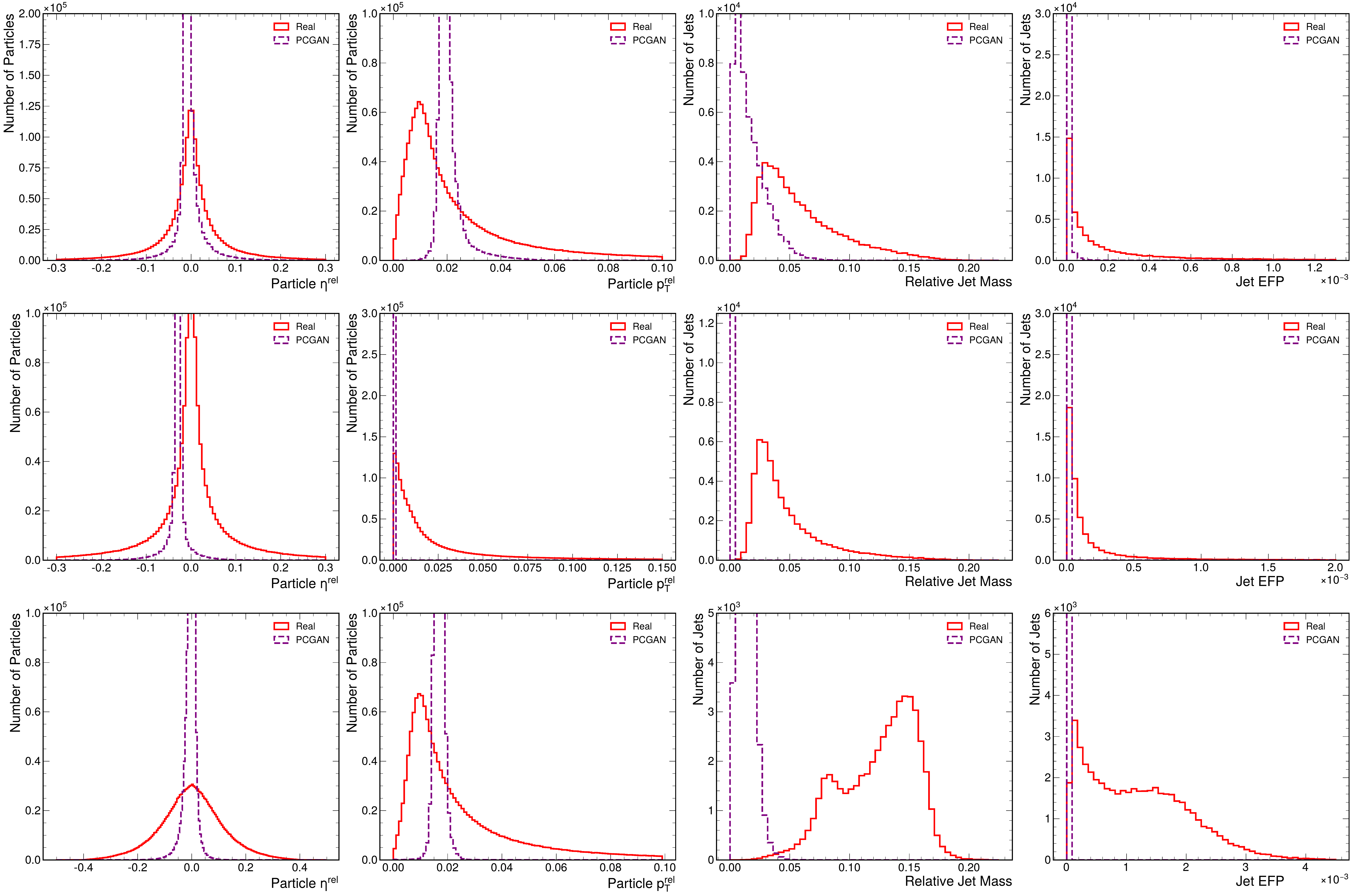}}
    \caption{Comparison of real and PCGAN-generated distributions for a subset of jet and particle features. Top: gluon jet features, Middle: light quark jets, Bottom: top quark jets.}
    \label{fig:pcganresults}
\end{figure}

Apart from the GAN models discussed in Sec.~\ref{sec:relatedwork}, there are several published generative models for point clouds which we argue are not applicable to jets.

\subsection{ShapeNet Point Clouds}
A number of successful generative models exploit a key inductive bias of ShapeNet-based clouds: that the individual distributions of sampled points conditioned on a particular object are identical and independent (the i.i.d assumption). 
This assumption allows for hierarchical generative frameworks, such as Point-Cloud-GAN (PCGAN)~\cite{pcgan}, which uses two networks: one to generate a latent object-level representation, and a second to sample independent points given such a representation.
The PointFlow~\cite{pointflow} and Discrete PointFlow~\cite{discretepointflow} models use a similar idea of sampling independently points conditioned on a learnt latent representation of the shape, but with a variational autoencoder (VAE) framework and using normalizing flows for transforming the sampled points. 

This hierarchical-sampling approach is appealing for ShapeNet clouds, however, as discussed in Sec.~\ref{sec:jets} the key i.i.d. assumption is not applicable to jets with their highly correlated particle constituents.
In fact, in contrast to ShapeNet objects which have a structure independent of the particular sampled cloud, jets are entirely defined by the distribution of their constituents.

Another model, ShapeGF~\cite{ShapeGF}, uses an approach of again sampling points independently from a prior distribution, but transforming them to areas of high density via gradient ascent, maximizing a learnt log-density concentrated on an object’s surface.
This approach suffers as well from the i.i.d. assumption in the context of jets, and additionally, unlike for ShapeNet point clouds, there is no such high-density region in momentum-space where particles tend to be concentrated, so learning and maximizing a log-density is not straightforward.

To support our overall claim of the inviability of the i.i.d. assumption for particle clouds, we train a PCGAN model on JetNet and show the produced feature distributions in Fig.~\ref{fig:pcganresults}. 
We can see, as expected, while this network is partially reproducing the particle feature distributions, it is entirely unable to learn the jet-level structure in particle clouds.

\subsection{Molecular Point Clouds}

3D molecules are another common point-cloud-style data structure, and there have been developments in generative models in this area as well.
Kohler et al.~\cite{kohler20} introduce physics-motivated normalizing flows equivariant to rotations around the center of mass, i.e. the SO(N) symmetries, for generating point clouds. 
This is appealing as normalizing flows give access to the explicit likelihood of generated samples, and having an architecture equivariant to physical symmetries such as 3D rotations can improve the generalizability and interpretability of the model. 
Since jets are relativistic, however, we require an architecture equivariant to the non-compact SO(3, 1) Lorentz group, to which this model has not been generalized yet. 
Simm et al.~\cite{simm21} present a reinforcement-learning-based approach for generating 3D molecules, using an agent to iteratively add atoms to a molecule and defining the reward function as the energy difference between the new molecule and the old with the new atom at the origin.
This reward function is not directly applicable to jets. where particle distributions are based on the QCD dynamics rather than on minimizing the total energy. 
Finally, Gebauer et al.~\cite{gschnet} introduce G-SchNet, an autoregressive model for producing molecules represented as point clouds, iteratively adding one atom at a time based on the existing molecule. 
Their iterative procedure however was proposed for point clouds of at most nine atoms, and does not scale well in terms of time to larger clouds.

Overall, all the models discussed heavily incorporate inductive biases which are specific to their respective datasets and don’t apply to JetNet.
However, they are extremely interesting approaches nonetheless, and adapting them with jet-motivated biases should certainly be explored in future work. 
Indeed, a significant contribution of our work is publishing a dataset which can facilitate and hopes to motivate such development.

\section{Training and Implementation Details}
\label{app:training}

PyTorch code and trained parameters for models in Table~\ref{tab:results} are provided in the MPGAN repository~\cite{mpgancode}. 
Models were trained and hyperparameters optimized on clusters of NVIDIA GeForce RTX 2080 Ti, Tesla V100, and A100 GPUs. 

\subsection{MPGAN}

We use the least squares loss function~\cite{LSGAN} and the RMSProp optimizer with a two time-scale update rule~\cite{TTUR} with a learning rate (LR) for the discriminator three times greater than that of the generator. The absolute rate differed per jet type. 
We use LeakyReLU activations (with negative slope coefficient 0.2) after all MLP layers except for the final generator and discriminator outputs where tanh and sigmoid activations respectively are applied. 
We attempted discriminator regularization to alleviate mode collapse via dropout~\cite{dropout}, batch normalization~\cite{batchnorm}, a gradient penalty~\cite{wgangp}, spectral normalization~\cite{spectralnorm}, adaptive competitive gradient descent~\cite{acgd} and data augmentation of real and generated graphs before the discriminator~\cite{karras_2020, tran_2020, zhao_2020}.
Apart from dropout (with fraction $0.5$), none of these demonstrated significant improvement with respect to mode dropping or cloud quality.

We use a generator LR of $10^{-3}$ and train for 2000 epochs for gluon jets, $2\times10^{-3}$ and 2000 epochs for top quark jets, and $0.5\times10^{-3}$ and 2500 epochs for light quark jets. 
We use a batch size of 256 for all jets. 

\subsection{rGAN, GraphCNNGAN, TreeGAN, and PointNet-Mix}

For rGAN and GraphCNNGAN we train two variants: (1) using the original architecture hyperparameters in Refs.~\cite{rgan, graphcnngan} for the 2048-node point clouds, and (2) using hyperparameters scaled down to 30-node clouds---specifically: a 32 dimensional latent space, followed by layers of 64, 128, and 90 nodes for r-GAN, or followed by two graph convolutional layers with node features sizes of 32 and 24 respectively for GraphCNN-GAN.
The scaled-down variant performed better for both models, and its scores are the ones reported in Table~\ref{tab:results}.
For TreeGAN, starting from single vertex---in analogy with a jet originating from a single particle---we use five layers of up-sampling and ancestor-descendant message passing, with a scale-factor of two in each and node features per layer of 96, 64, 64, 64, and 64 respectively.  
LRs, batch sizes, loss functions, gradient penalties, optimizers, ratios of critic to generator updates, activations, and number of epochs are the same as in the original paper and code. 
We use the architecture defined in~\cite{wang2020rethinking} for the PointNet-Mix discriminator. 

\subsection{FPND}
\label{app:pnet}

Apart from the number of input particle features (three in our case, excluding the mask feature), we use the original ParticleNet architecture in Ref.~\cite{particlenet}. 
We find training with the Adam optimizer, LR $10^{-4}$, for 30 epochs outperformed the original recommendations on our dataset. 
Activations after the first fully connected layer, pre-ReLU, are used for the FPND measurement.

\subsection{PCGAN}
\label{app:pcgan}

We use the original PCGAN implementation for the sampling networks and training, with a 256-dimensional latent object representation.
For the latent code GAN we use a 3 layer fully connected network for both the generator, with an input size of 128 and intermediate layer sizes of 256 and 512, and discriminator, with intermediate layer sizes of 512 and 256, trained using the Wasserstein-GAN~\cite{WGAN} loss with a gradient penalty.

% \section{Locally Connected Graphs}
% \label{app:locallyconnected}

% \TODO{Describe non-convergence as knn is lowered}

\section{Masking Strategies}
\label{app:masking}

% \TODO{Clean this up.}

\begin{figure}[t]
    \centering
    \centerline{\includegraphics[width=\textwidth]{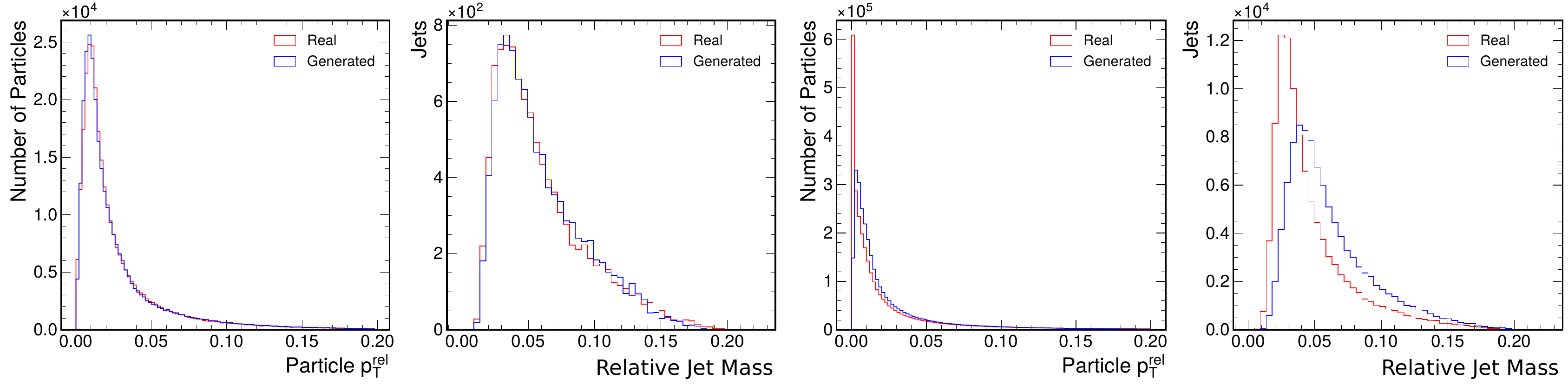}}
    \caption{Particle $\ptrel$ and relative jet mass distributions of real jets and those generated by MPGAN without our masking strategy. Left: gluon, right: light quark jets.
    We see that while for gluon jets the generator learns distributions correctly, it struggles to learn the discontinuous spike, due to the zero-padded particles, in the light quark $\ptrel$ distribution.
    This also leads to a distorted mass distribution.
    }
    \label{fig:zeropadding}
\end{figure}

In JetNet, jets with fewer than 30 particles are zero-padded to fill the 30-particle point cloud. 
Such zero-padded particles pose a problem for the generator, which is not able to learn this sharp discontinuity in the jet constituents (Fig.~\ref{fig:zeropadding}).

\begin{figure}[htpb]
    \centering
    \centerline{\includegraphics[width=\textwidth]{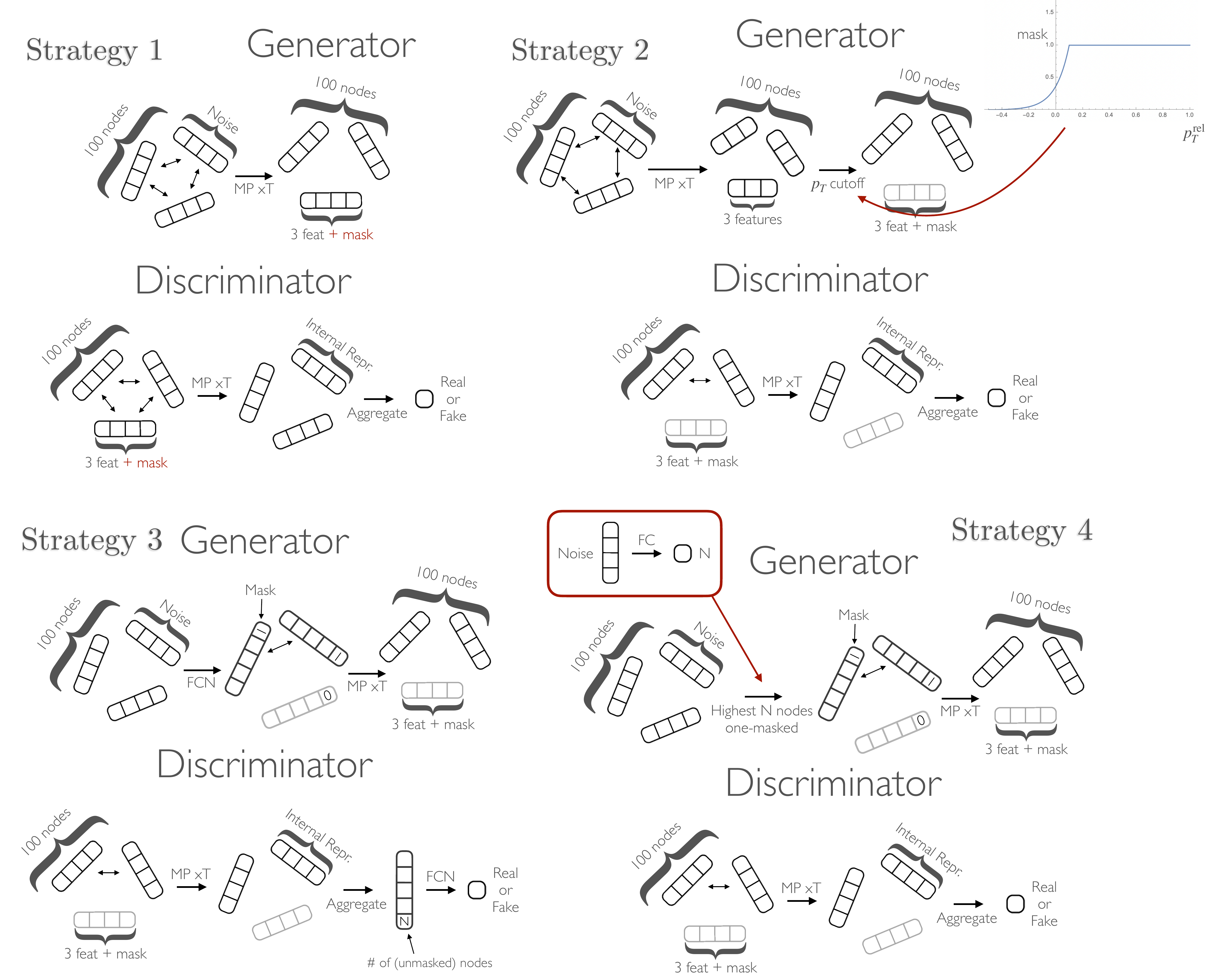}}
    \caption{The four alternative masking strategies which we test.
    }
    \label{fig:masking}
\end{figure}

\begin{figure}[htpb]
    \centering
    \centerline{\includegraphics[width=0.6\textwidth]{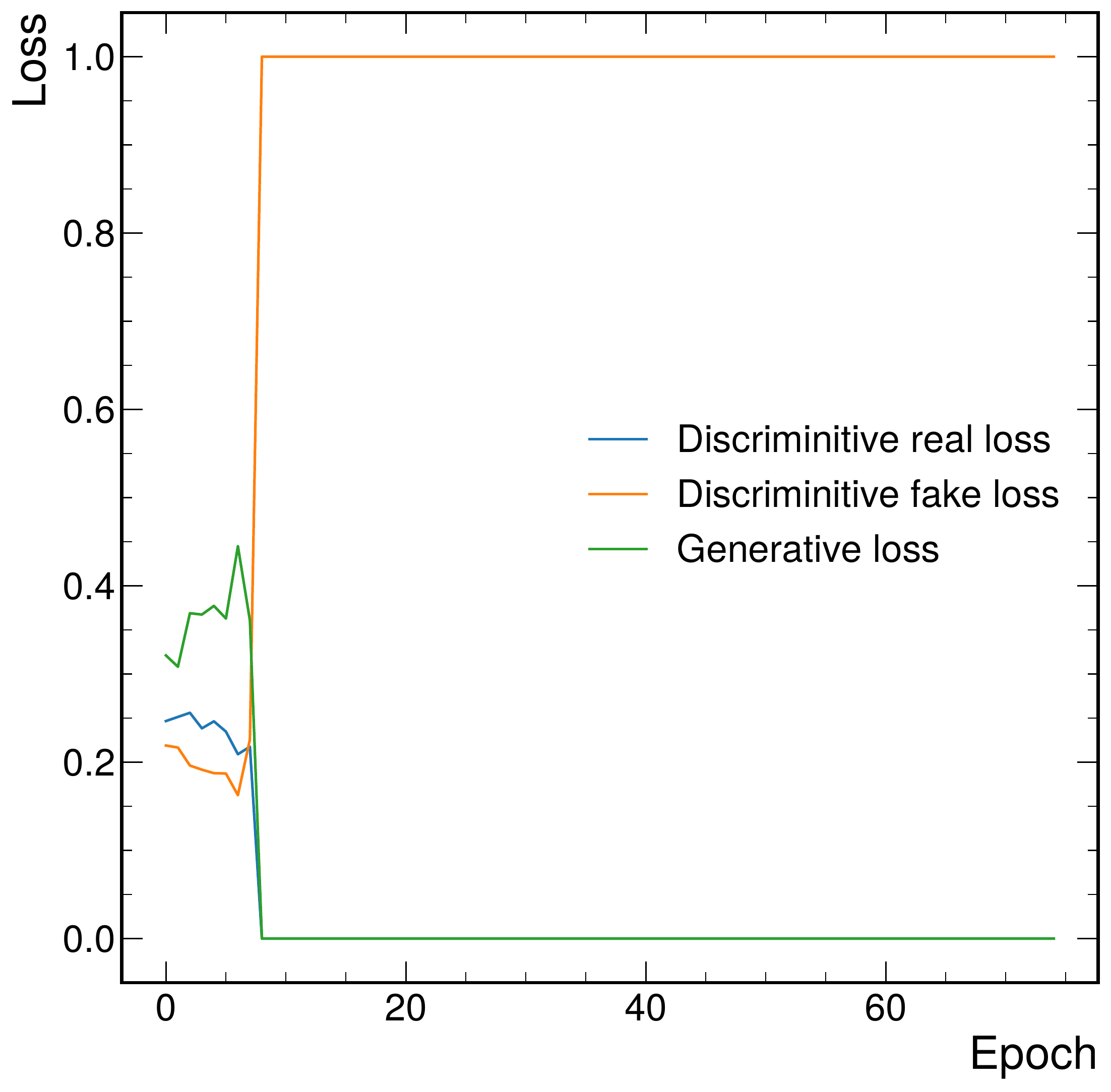}}
    \caption{Loss curve of a training on light quark jets with masking strategy 3, typical of loss curves with all four strategies.
    }
    \label{fig:masking_loss}
\end{figure}

To counter this issue, we experiment with five masking strategies, out of which the one described in Sec.~\ref{sec:arch} was most successful. 
The four alternatives, which all involve the generator learning the mask without any external input, are shown in Fig.~\ref{fig:masking}.

Strategy 1 treats the mask homogeneously as an extra feature to learn. 
A variation of this weights the nodes in the discriminator the mask. 
In strategy 2, a mask is calculated for each generated particle as a function of its $\ptrel$, based on an empirical minimum cutoff in the dataset. 
In particular, both a Heaviside-step-function and a continuous mask function as in the figure are tested. 
The standard MP discriminator, as described in Sec.~\ref{sec:arch}, is used. 
Strategy 3 sees the generator applying an FC layer per particle in the initial cloud to learn their respective masks, with both the MP discriminator, as well as a variant with the number of unmasked nodes in the clouds added as an extra feature to the FC layer. 
In strategies 1 and 3 we test
both binary and continuous masks.
Finally, in strategy 4, we train an auxiliary network to choose a number of particles to mask (as opposed to sampling from the real distribution), which is then passed into the standard MP generator.

We find that all such strategies are unable to produce accurate light quark jets, and in fact trainings for each diverge in the fashion depicted in Fig.~\ref{fig:masking_loss}, even using each discriminator regularization method mentioned in App.~\ref{app:training}).
We conclude that learning the number of particles to produce is a significant challenge for a generator network, but is a relatively simple feature with which to discriminate between real and fake jets.
To equalize this we use the strategy in Sec.~\ref{sec:arch} where the number of particles to produce is sampled directly from the real distribution, removing the burden of learning this distribution from the generator network.

\section{Jet Images}
\label{app:jetimages}

\begin{figure}[htpb]
    \centering
    \centerline{\includegraphics[width=1.2\textwidth]{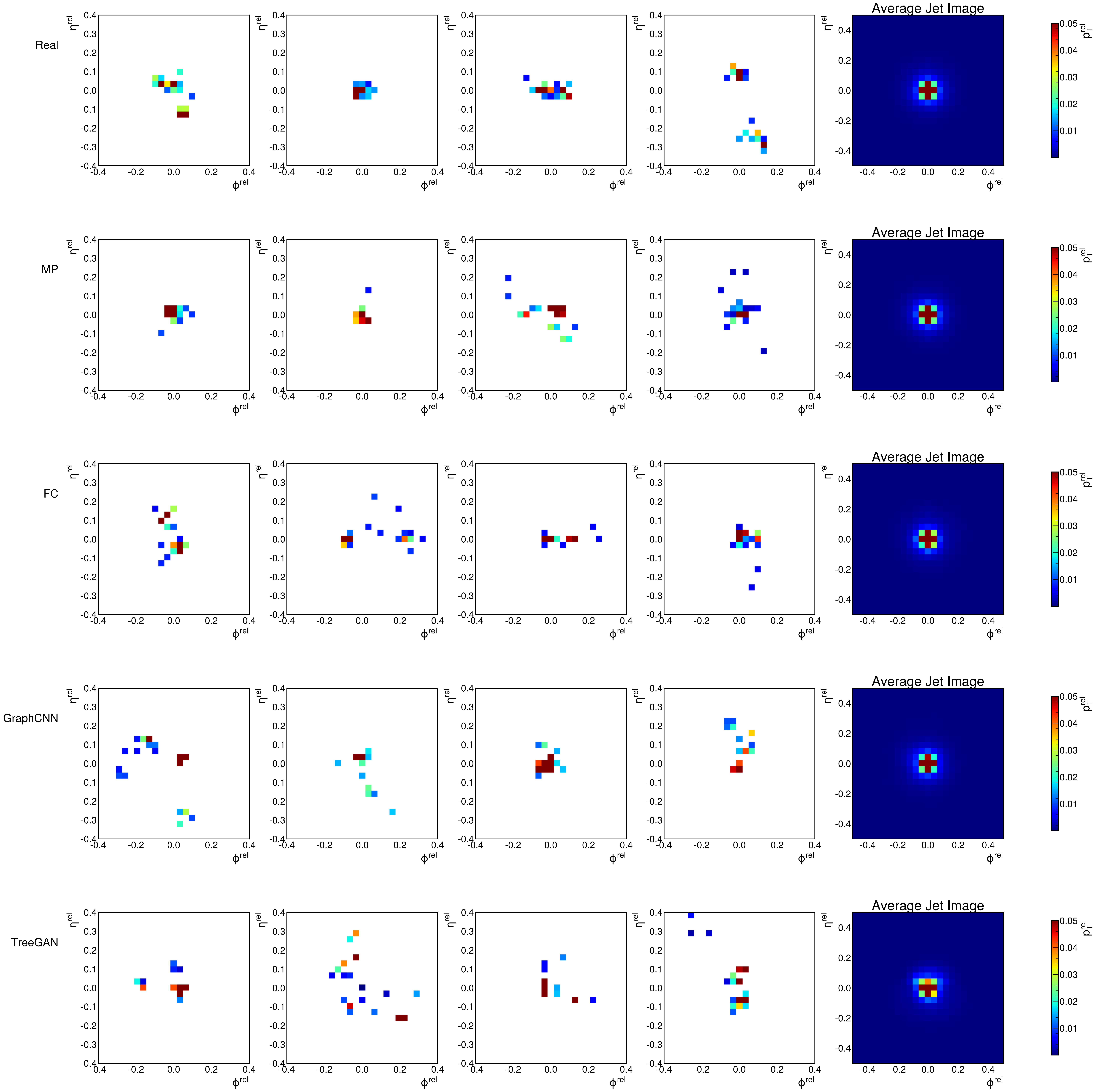}}
    \caption{Random samples of discretized images in the \etarel$-$\phirel plane, with pixel intensities equal to particle \ptrel, of real and generated gluon jets (left), and an average over 10,000 such sample images (right).}
    \label{fig:jetims_g}
\end{figure}

\begin{figure}[htpb]
    \centering
    \centerline{\includegraphics[width=1.2\textwidth]{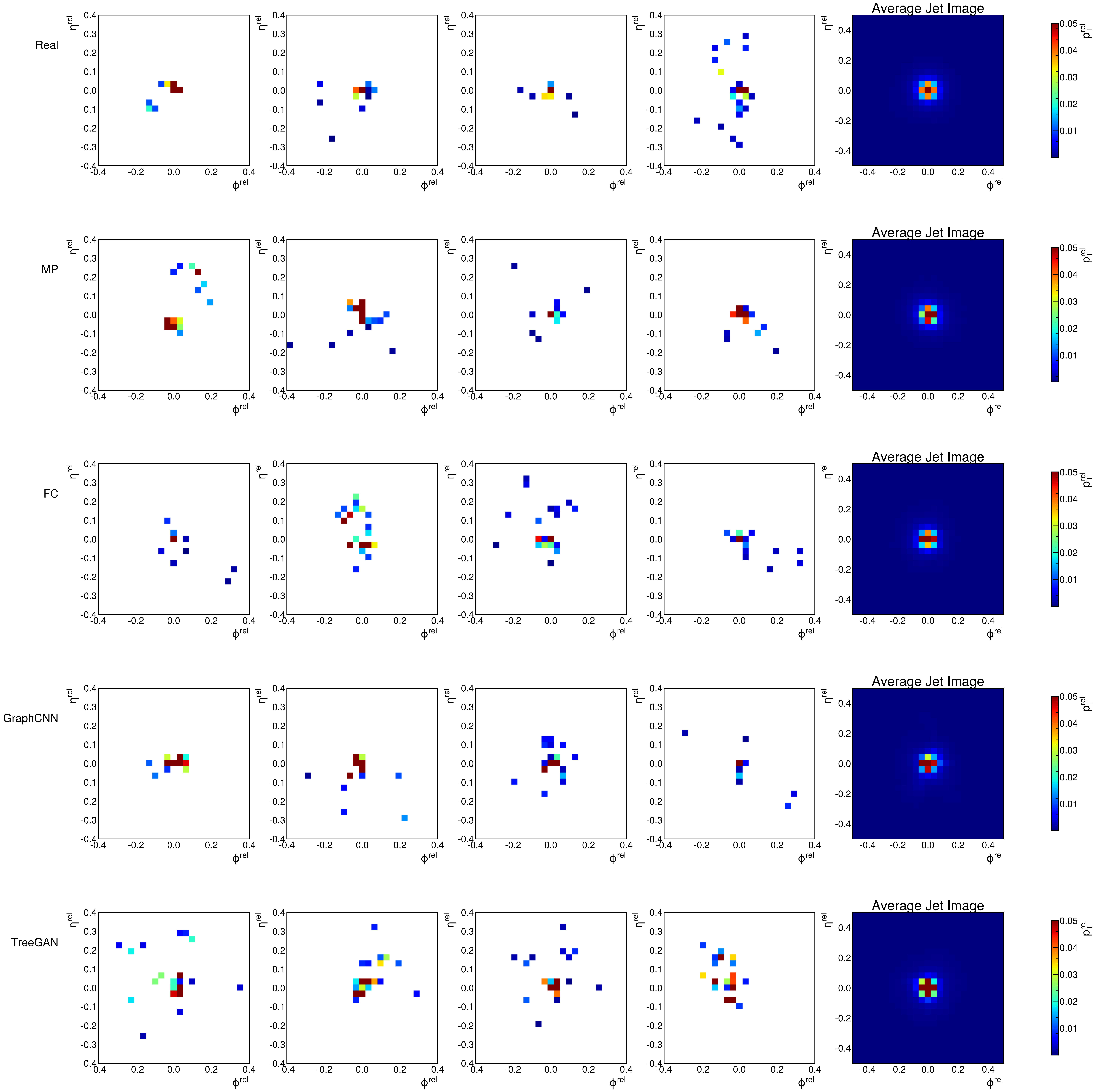}}
    \caption{Random samples of discretized images in the \etarel$-$\phirel plane, with pixel intensities equal to particle \ptrel, of real and generated light quark jets (left), and an average over 10,000 such sample images (right).}
    \label{fig:jetims_q}
\end{figure}

\begin{figure}[htpb]
    \centering
    \centerline{\includegraphics[width=1.2\textwidth]{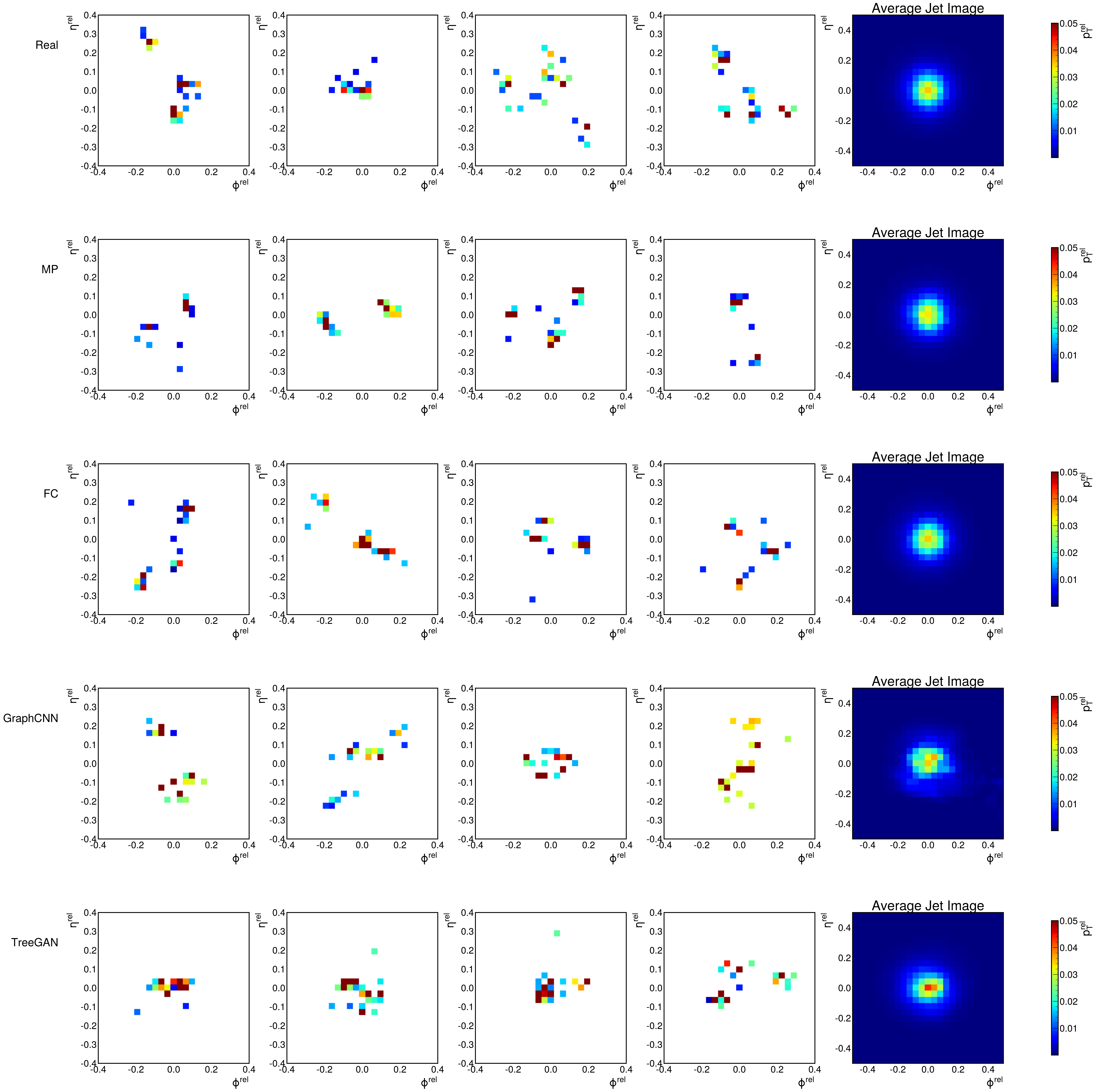}}
    \caption{Random samples of discretized images in the \etarel$-$\phirel plane, with pixel intensities equal to particle \ptrel, of real and generated top quark jets (left), and an average over 10,000 such sample images (right).}
    \label{fig:jetims_t}
\end{figure}

Figs.~\crefrange{fig:jetims_g}{fig:jetims_t} show samples of real and generated ``jet images'': discretized representations of jets in the angular-coordinate-plane.

\end{document}